\documentclass[10pt,twocolumn,letterpaper]{article}

\usepackage{cvpr}
\usepackage{times}
\usepackage{epsfig}
\usepackage{graphicx}
\usepackage{amsmath}
\usepackage{amssymb}

\usepackage{enumitem}
\usepackage{dsfont}
\newcommand{\vect}[1]{\boldsymbol{#1}}
\usepackage{subcaption}
\usepackage[ruled,vlined,linesnumbered]{algorithm2e}
\usepackage{pifont}
\newcommand{\cmark}{\ding{51}}%
\newcommand{\xmark}{\ding{55}}%

\usepackage[pagebackref=true,breaklinks=true,letterpaper=true,colorlinks,bookmarks=false]{hyperref}

\cvprfinalcopy 

\def\cvprPaperID{2712} 

\ifcvprfinal\fi
\begin{document}

\title{Rethinking Normalization and Elimination Singularity in Neural Networks}

\author{Siyuan Qiao~~~~Huiyu Wang~~~~Chenxi Liu~~~~Wei Shen~~~~Alan Yuille\\
Johns Hopkins University\\
{\tt\small \{siyuan.qiao, hwang157, cxliu\}@jhu.edu~~~~\{shenwei1231, alan.l.yuille\}@gmail.com}
}

\maketitle

\begin{abstract}

In this paper, we study normalization methods for neural networks from the perspective of elimination singularity.
Elimination singularities correspond to the points on the training trajectory where neurons become consistently deactivated.
They cause degenerate manifolds in the loss landscape which will slow down training and harm model performances.
We show that channel-based normalizations (\textit{e.g.} Layer Normalization and Group Normalization) are unable to guarantee a far distance from elimination singularities, in contrast with Batch Normalization which by design avoids models from getting too close to them.
To address this issue, we propose Batch-Channel Normalization (BCN), which uses batch knowledge to avoid the elimination singularities in the training of channel-normalized models.
Unlike Batch Normalization, BCN is able to run in both large-batch and micro-batch training settings.
The effectiveness of BCN is verified on many tasks, including image classification, object detection, instance segmentation, and semantic segmentation.
The code is here: {\small\url{https://github.com/joe-siyuan-qiao/Batch-Channel-Normalization}.}

\end{abstract}

\vspace{-0.2in}

\section{Introduction}

Deep neural networks achieve state-of-the-art results in many vision tasks~\cite{deeplab,maskrcnn,resnet}.
Despite being very effective, deep networks are hard to train.
Normalization methods~\cite{layernorm,batchnorm} are crucial for stabilizing and accelerating network training.
There are many theories explaining how normalizations help optimization.
For example, Batch Normalization (BN)~\cite{batchnorm} and Layer Normalization (LN)~\cite{layernorm} were proposed based on the conjecture that they are able to reduce internal covariate shift which negatively impacts training.
Santurkar \textit{et al.}~\cite{whybnworks} argue that the reason of the success of BN is that it makes the loss landscape significantly smoother.
Unlike the previous work, we study normalizations from the perspective of avoiding elimination singularities~\cite{wei2008dynamics} that also have negative effects on training.

Elimination singularities refer to the points along the training trajectory where neurons in the networks get eliminated.
As shown in \cite{orhan2017skip}, the performance of neural networks is correlated with their distance to elimination singularities:
{\it the closer the model is to the elimination singularities, the worse it performs}.
Sec.~\ref{sec:elim} provides a closer look at the relationship between the performance and the distance through experiments.
Because of this relationship, we ask:
\begin{quote}\it
    Do all the normalization methods keep their models away from elimination singularities?
\end{quote}
Here, we list our findings:
\begin{enumerate}[topsep=0.5ex,itemsep=-0.5ex,partopsep=0.5ex,parsep=0.5ex,leftmargin=0.6cm,label*=(\arabic*)]
    \item Batch Normalization (BN)~\cite{batchnorm} is \textbf{able} to keep models at far distances from the singularities.
    \item Channel-based normalization, \textit{e.g.,} Layer Normalization (LN)~\cite{layernorm} and Group Normalization (GN)~\cite{groupnorm}, is \textbf{unable} to guarantee far distances, where the situation of LN is worse than that of GN.
    \item Weight Standardization (WS)~\cite{qiao2019weight,huang2017centered} is able to push the models away from the elimination singularities.
\end{enumerate}
These findings provide a new way of understanding why GN is performing better than LN, and how WS improves the performances of both of them.

Since channel-based normalization methods (\textit{e.g.} LN and GN) have issues with elimination singularity, we can improve their performances if we are able to push models away from them.
For this purpose, we propose Batch-Channel Normalization (BCN), which uses batch knowledge to prevent channel-normalized models from getting too close to the elimination singularities.
Sec.~\ref{sec:norm} shows the detailed modeling of the proposed normalization method.
Unlike BN, BCN is able to run in both \textbf{large-batch} and \textbf{micro-batch} training settings and can improve the performances of channel-normalized models.

To evaluate our proposed BCN, we test it on various popular vision tasks, including large-batch training of ResNet~\cite{resnet} on ImageNet~\cite{ILSVRC15}, large-batch training of DeepLabV3~\cite{deeplabv3} on PASCAL VOC~\cite{pascal}, and micro-batch training of Faster R-CNN~\cite{fasterrcnn} and Mask R-CNN~\cite{maskrcnn} on MS COCO~\cite{coco} dataset.
Sec.~\ref{sec:exp} shows the experimental results, which demonstrate that our proposed BCN is able to outperform the baselines effortlessly.
Finally, Sec.~\ref{sec:related} discusses the related work, and Sec.~\ref{sec:conc} concludes the paper.

\section{Normalization and Elimination Singularity}\label{sec:elim}

In this section, we will provide the background of normalization methods, discuss the relationship between the performance and the distance to elimination singularities, and show how well normalization methods are able to prevent models from getting too close to those singularities.

\subsection{Batch- and Channel-based Normalization}
Based on how activations are normalized, we group the normalization methods into two types: batch-based normalization and channel-based normalization, where the batch-based normalization method corresponds to BN and the channel-based normalization methods include LN and GN.

Suppose we are going to normalize a 2D feature map $\vect{X}\in\mathds{R}^{B\times C\times H\times W}$, where $B$ is the batch size, $C$ is the number of channels, $H$ and $W$ denote the height and the width.
For each channel $c$, BN normalizes $\vect{X}$ by
\begin{equation}\label{eq:bn}
    \vect{Y}_{\cdot c \cdot\cdot} = \dfrac{\vect{X}_{\cdot c\cdot\cdot} - \mu_{\cdot c\cdot\cdot}}{\sigma_{\cdot c\cdot\cdot}}
\end{equation}
where $\mu_{\cdot c\cdot\cdot}$ and $\sigma_{\cdot c\cdot\cdot}$ denote the mean and the standard deviation of all the features of the channel $c$, $\vect{X}_{\cdot c\cdot\cdot}$.
Throughout the paper, we use $\cdot$ in the subscript to denote all the features along that dimension for convenience.

Unlike BN which computes statistics on the batch dimension in addition to the height and width, channel-based normalization methods compute statistics on the channel dimension.
Specifically, they divide the channels to several groups, and normalize each group of channels together, \textit{i.e.},
$\vect{X}$ is reshaped as $\vect{\dot{X}}\in\mathds{R}^{B\times G\times C/G\times H\times W}$, and then:
\begin{equation}\label{eq:cn}
    \vect{\dot{Y}}_{bg\cdot\cdot\cdot} = \dfrac{\vect{\dot{X}}_{bg\cdot\cdot\cdot} - \mu_{bg\cdot\cdot\cdot}}{\sigma{_{bg\cdot\cdot\cdot}}}
\end{equation}
for each sample $b$ of $B$ samples in a batch and each channel group $g$ out of all $G$ groups.
After Eq.~\ref{eq:cn}, the output $\vect{\dot{Y}}$ is reshaped as $\vect{\dot{X}}$ and denoted by $\vect{Y}$.

Both batch- and channel-based normalization methods optionally have an affine transformation, \textit{i.e.},
\begin{equation}\label{eq:at}
    \vect{Z}_{\cdot c\cdot\cdot} = \gamma_c\vect{Y}_{\cdot c\cdot\cdot} + \beta_c
\end{equation}

\subsection{Performance and Distance to Singularities}\label{sec:dist}

Deep neural networks are hard to train partly due to the singularities caused by the non-identifiability of the model~\cite{wei2008dynamics}.
These singularities include overlap singularities, linear dependence singularities, elimination singularities, \textit{etc}.
Degenerate manifolds in the loss landscape will be caused by these singularities, getting closer to which will slow down learning and impact model performances~\cite{orhan2017skip}.
In this paper, we focus on elimination singularities, which correspond to the points on the training trajectory where neurons in the model become constantly deactivated.

We focus on a basic building element that is widely used in neural networks: a convolutional layer followed by a normalization method (\textit{e.g.} BN, LN) and ReLU~\cite{relu}, \textit{i.e.},
\begin{equation}
   \vect{X^{\text{out}}} = \text{ReLU}(\text{Norm}(\text{Conv}(\vect{X^{\text{in}}})))
\end{equation}
ReLU sets any values below $0$ to $0$, thus a neuron is constantly deactivated if its maximum value after normalization is below $0$.
Their gradients will also be $0$ because of ReLU, making them hard to revive; hence, a singularity is created.

\paragraph{BN avoids elimination singularities.}
Here, we study the effect of BN on elimination singularities.
Since the normalization methods all have an optional affine transformation, we focus on the distinct part of BN: Eq.~\ref{eq:bn}, which normalizes all channels to zero mean and unit variance, \textit{i.e.},
\begin{equation}\label{eq:bnes}
    \mathbb{E}_{y\in Y_{\cdot c\cdot\cdot}}\big[y\big]=0,~~ \mathbb{E}_{y\in Y_{\cdot c\cdot\cdot}}\big[y^2\big]=1,~~\forall c
\end{equation}
As a result, regardless of the weights and the distribution of the inputs, Eq.~\ref{eq:bn} guarantees that the activations of each channel are zero-centered with unit variance.
Therefore, each channel cannot be constantly deactivated because there are always some activations that are $>0$, nor almost constantly deactivated caused by one channel having a very small activation scale compared with the others.

\paragraph{Statistics affect the distance and the performance.}

\begin{figure}
    \centering
    \includegraphics[width=\linewidth]{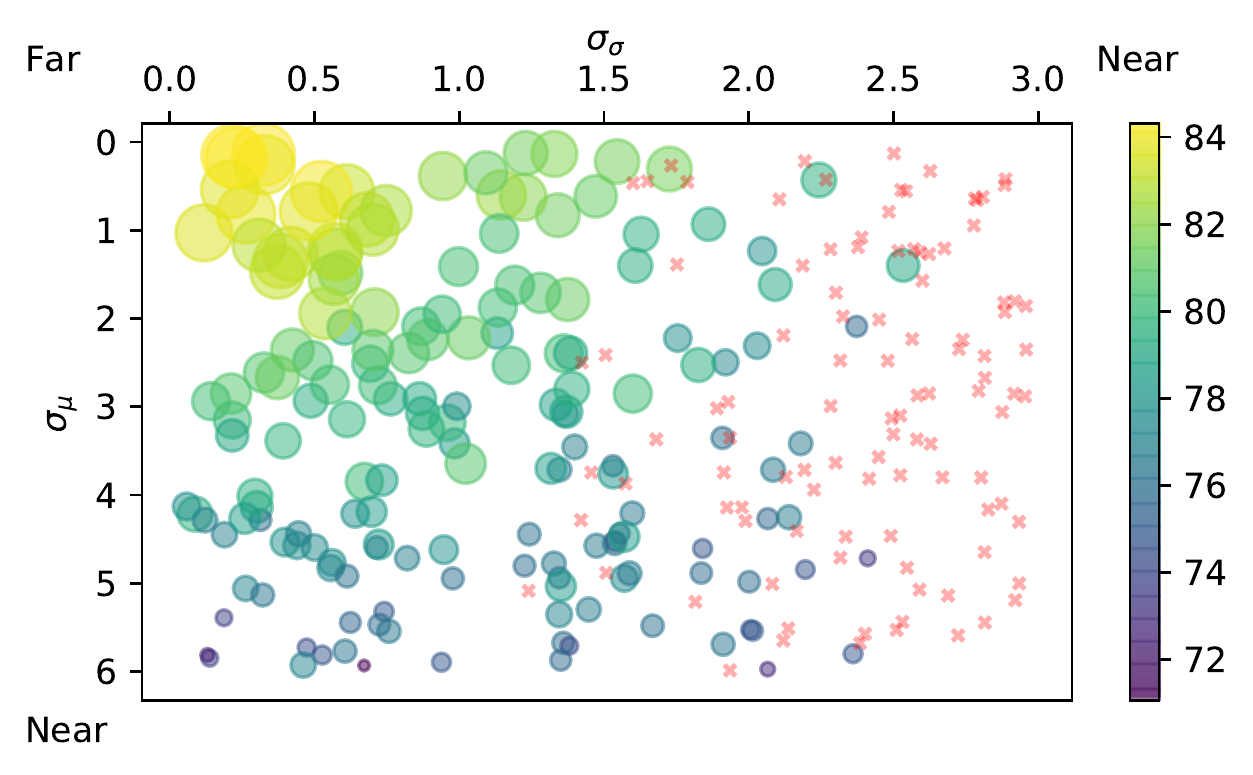}
    \caption{Model accuracy and distance to singularities.
    Larger circles correspond to higher performances.
    Red crosses represent failure cases (accuracy $<70\%$).
    Circles are farther from singularities if they are closer to the origin.}
    \label{fig:dist}
\end{figure}

\begingroup
\advance\leftmargini -1em
\begin{quote}
\it
    BN avoids singularities by normalizing each channel to zero mean and unit variance.
    What if they are normalized to other means and variances?
\end{quote}
\endgroup

We ask this question because this is similar to what happens in channel-normalized models.
Channel-based normalization methods, as they do \textbf{not} have batch information, are \textbf{unable} to make sure all neurons have zero mean and unit variance after normalization.
Instead, they will have \textbf{different} statistics, thus make the model closer to singularities.
Here, by \textit{closer}, we mean the model is \textit{far} from BN where each channel is zero-centered with unit variance, which avoids all singularities.
To study the relationship between the performance and the distance to singularities (or how far from BN) caused by statistical differences, we conduct experiments on a 4-layer convolutional network.
Each convolutional layer has 32 output channels, and is followed by an average pooling layer which down-samples the features by a factor of 2.
Finally, a global average pooling layer and a fully-connected layer will output the logits for Softmax.
The experiments are done on CIFAR-10~\cite{cifar}.

In the experiment, each channel $c$ will be normalized to a pre-defined mean $\hat{\mu}_{c}$ and a pre-defined variance $\hat{\sigma}_{c}$ that are drawn from two distributions, respectively:
\begin{equation}
    \hat{\mu}_{c}\sim\mathcal{N}(0, \sigma_{\mu}) ~~\text{and}~~ \hat{\sigma}_{c}=e^{\dot{\sigma}_{c}}~\text{where}~ \dot{\sigma}_{c}\sim\mathcal{N}(0,\sigma_\sigma)
\end{equation}
\textit{The model will be closer to singularities when $\sigma_{\mu}$ or $\sigma_{\sigma}$ increases.
BN corresponds to the case where $\sigma_{\mu}=\sigma_{\sigma}=0$}.

After getting $\hat{\mu}_{c}$ and $\hat{\sigma}_{c}$ for each channel, we compute
\begin{equation}
    \vect{Y}_{\cdot c\cdot\cdot} = \gamma_c\big(\hat{\sigma}_{c}\dfrac{\vect{X}_{\cdot c\cdot\cdot} - \mu_{\cdot c\cdot\cdot}}{\sigma_{\cdot c\cdot\cdot}} + \hat{\mu}_{c}\big) + \beta_c
\end{equation}
Note that $\hat{\mu}_{c}$ and $\hat{\sigma}_{c}$ are fixed during training while $\gamma_c$ and $\beta_c$ are trainable parameters in the affine transformation.

\begin{figure}
    \centering
    \begin{subfigure}[b]{\linewidth}
        \centering
         \includegraphics[width=\linewidth]{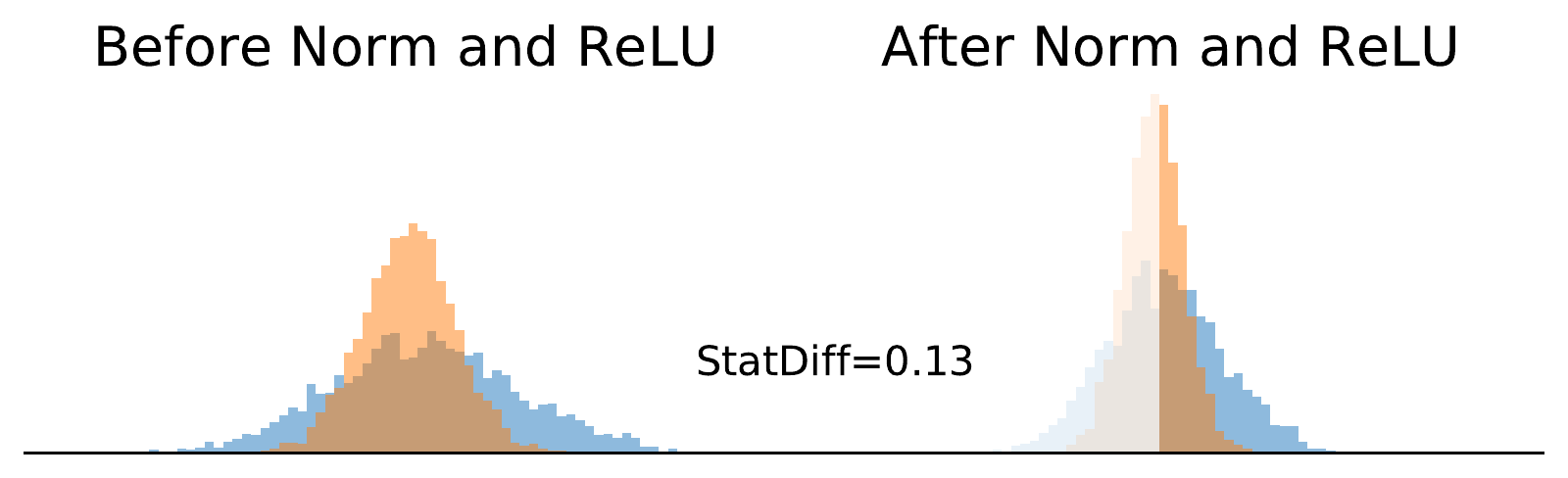}
    \end{subfigure}
    \begin{subfigure}[b]{\linewidth}
        \centering
         \includegraphics[width=\linewidth]{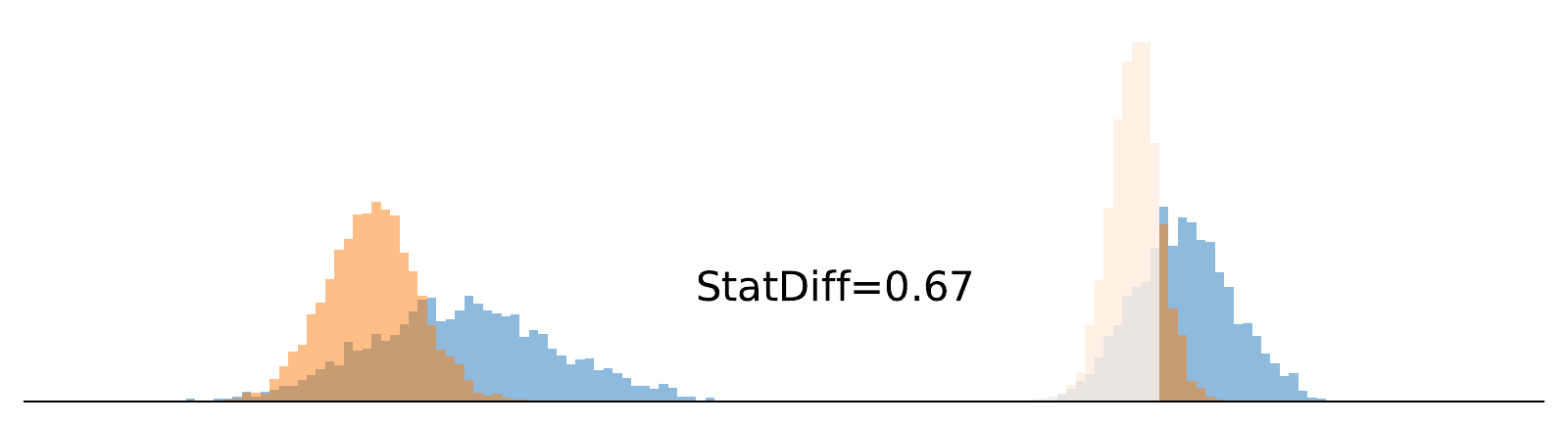}
    \end{subfigure}
    \begin{subfigure}[b]{\linewidth}
        \centering
         \includegraphics[width=\linewidth]{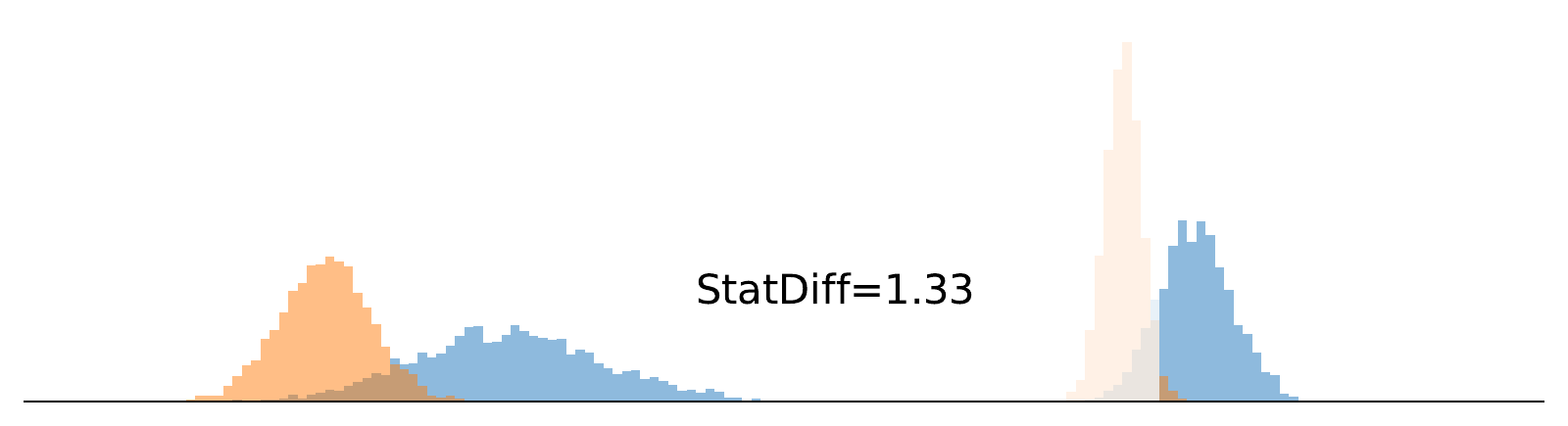}
    \end{subfigure}
    \caption{Examples of normalizing two channels in a group when they have different means and variances.
    Transparent bars mean they are $0$ after ReLU.
    StatDiff defined in Eq.~\ref{eq:sd}.}
    \label{fig:dist_exp}
\end{figure}

Fig.~\ref{fig:dist} shows the experimental results.
When $\sigma_{\mu}$ and $\sigma_{\sigma}$ are closer to the origin, the normalization method is more similar to BN, and the model will be farther from the singularities.
When their values increase, we observe performance decreases.
For extreme cases, we also observe training failures.
These results indicate that although the affine transformation theoretically can find solutions that cancel the negative effects of normalizing channels to different statistics, their capability is limited by the gradient-based training.
These findings raise concerns about channel normalizations regarding their distance to singularities.

\subsection{Statistics in Channel Normalization}
Following our concerns about channel-based normalization and their distance to singularities, we study the statistical differences between channels when they are normalized by a channel-based normalization such as GN or LN.

\begin{figure}
    \centering
    \includegraphics[width=\linewidth]{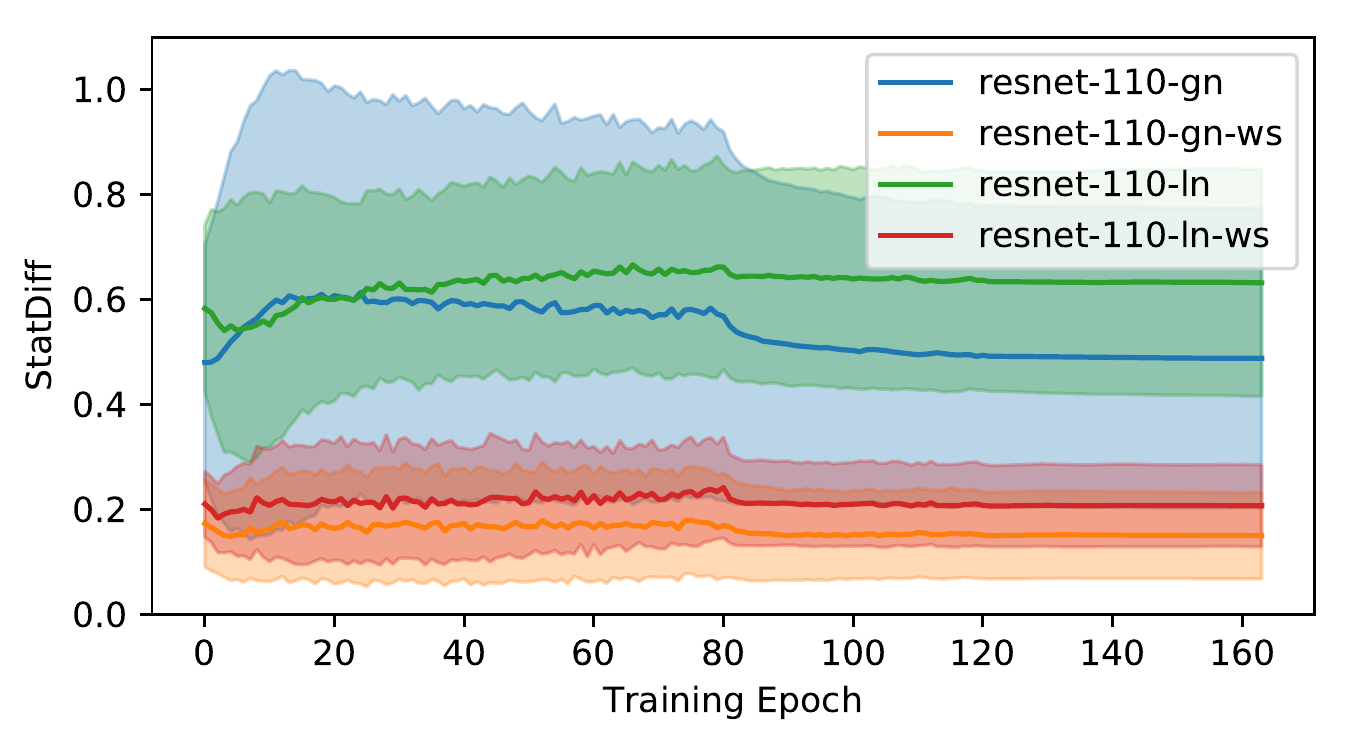}
    \caption{Means and standard deviations of the statistical differences (StatDiff in Eq.~\ref{eq:sd}) of all layers in a ResNet-110 trained on CIFAR-10 with GN, GN+WS, LN, and LN+WS.}
    \label{fig:stat_diff}
\end{figure}

\paragraph{Statistical differences in GN, LN and WS.}
We train a ResNet-110~\cite{resnet} on CIFAR-10~\cite{cifar} normalized by GN, LN, with and without WS~\cite{qiao2019weight}.
During training, we keep record of the running mean $\mu^r_c$ and variance $\sigma_c^r$ of each channel $c$ after convolutional layers.
For each group $g$ of the channels that are normalized together, we compute their channel \textbf{statistical difference} defined as the standard deviation of their means divided by the mean of their standard deviations, \textit{i.e.},
\begin{equation}\label{eq:sd}
    \text{StatDiff}(g) = \dfrac{ \sqrt{\mathbb{E}_{c\in g}\big[(\mu^r_{c})^2\big] - \big(\mathbb{E}_{c\in g}\big[\mu^r_{c}\big]\big)^2}}{\mathbb{E}_{c\in g}\big[ \sigma_{c} \big]}
\end{equation}
We plot the average statistical differences of all the groups after every training epoch as shown in Fig.~\ref{fig:stat_diff}.

By Eq.~\ref{eq:sd}, $\text{StatDiff}(g)\geq 0,~\forall g$.
In BN, all their means are the same, as well as their variances, thus $\text{StatDiff}(g)=0$.
As the value of $\text{StatDiff}(g)$ goes up, the differences between channels within a group become larger.
Since they will be normalized together as in Eq.~\ref{eq:cn}, large differences will inevitably lead to underrepresented channels.
Fig.~\ref{fig:dist_exp} plots 3 examples of 2 channels before and after normalization in Eq.~\ref{eq:cn}.
Compared with those examples, it is clear that the models in Fig.~\ref{fig:stat_diff} have many underrepresented channels.

\paragraph{Why GN performs better than LN.}
Fig.~\ref{fig:stat_diff} also provides explanations why GN performs better than LN.
Comparing GN and LN, the major difference is their numbers of groups for channels: LN has only one group for all the channels in a layer while GN collects them into several groups.
A strong benefit of having more than one group is that it guarantees that each group will at least have one neuron that is not suppressed by the others from the same group.
Therefore, GN provides a mechanism to prevent the models from getting too close to the elimination singularities.

Fig.~\ref{fig:stat_diff} also shows the statistical differences when WS is used.
From the results, we can clearly see that WS makes StatDiff much closer to $0$.
Consequently, the majority of the channels are not underrepresented in WS: most of them are frequently activated and they are at similar activation scales.
This makes training with WS easier and their results better.

\paragraph{Why WS helps.}
Here, we also provide our understandings why WS is able to achieve smaller statistical differences.
Recall that WS adds constraints to the weight $\vect{W}\in\mathds{R}^{\text{O}\times\text{I}}$ of a convolutional layer with O output channels and I inputs such that $\forall c$,
\begin{equation}\label{eq:ws1}
    \sum_{i=1}^I\vect{W}_{c,i} = 0,~~~\sum_{i=1}^I\vect{W}^2_{c,i} = 1
\end{equation}
With the constraints of WS, $\mu_{c}^{\text{out}}$ and $\sigma_{c}^{\text{out}}$ become
\begin{equation}\label{eq:ws2}
    \mu_{c}^{\text{out}}=\sum_{i=1}^I\vect{W}_{c,i}\mu_i^{\text{in}},~~~(\sigma_{c}^{\text{out}})^2=\sum_{i=1}^I\vect{W}_{c,i}^2(\sigma_i^{\text{in}})^2
\end{equation}
when we follow the assumptions in Xavier initialization~\cite{glorot2010understanding}.
When the input channels are similar in their statistics, \textit{i.e.}, $\mu_{i}^{\text{in}}\approx\mu_{j}^{\text{in}}$, $\sigma_{i}^{\text{in}}\approx\sigma_{j}^{\text{in}}$, $\forall i,j$,
\begin{eqnarray}
    \mu_{c}^{\text{out}}&\approx&\mu_1^{\text{in}}\sum_{i=1}^I\vect{W}_{c,i}=0 \\
    (\sigma_{c}^{\text{out}})^2&\approx&(\sigma_{1}^{\text{in}})^2\sum_{i=1}^I\vect{W}_{c,i}^2=(\sigma_{1}^{\text{in}})^2
\end{eqnarray}
In other words, WS can pass the statistical similarities from the input channels to the output channels, all the way from the image space where RGB channels are properly normalized.
This is similar to the objective of Xavier initialization~\cite{glorot2010understanding} or Kaiming initialization~\cite{he2015delving}, except that WS enforces it by reparameterization throughout the entire training process, thus is able to reduce the statistical differences.

Here, we summarize this subsection.
We have shown that channel-based normalization methods, as they do not have batch information, are not able to ensure a far distance from elimination singularities.
Without the help of batch information, GN alleviates this issue by assigning channels to more than one group to encourage more activated neurons, and WS adds constraints to pull the channels to be not so statistically different.
We notice that the batch information is not hard to collect in reality.
This inspires us to equip channel-based normalization with batch information, and the result is Batch-Channel Normalization.

\section{Batch-Channel Normalization}\label{sec:norm}

This section presents the definition of Batch-Channel Normalization, discusses why adding batch statistics to channel normalization is not redundant, and shows how BCN runs in large-batch and micro-batch training settings.

\subsection{Definition}
Batch-Channel Normalization (BCN) adds batch constraints to channel-based normalization methods.
Let $\vect{X}\in\mathds{R}^{B\times C\times H\times W}$ be the features to be normalized.
Then, the normalization is done as follows.
$\forall c$,
\begin{equation}\label{eq:bcnbn}
    \vect{\dot{X}}_{\cdot c\cdot\cdot}=\gamma_c^b \dfrac{\vect{X}_{\cdot c\cdot\cdot} - \hat{\mu}_{ c}}{\hat{\sigma}_{c}}+\beta_{c}^b
\end{equation}
where the purpose of $\hat{\mu}_{c}$ and $\hat{\sigma}_{c}$ is to make
\begin{equation}\label{eq:bcn_purpose}
    \mathbb{E}\Big\{\dfrac{\vect{X}_{\cdot c\cdot\cdot} - \hat{\mu}_{ c}}{\hat{\sigma}_{c}}\Big\}= 0~\text{and}~\mathbb{E}\Big\{\big(\dfrac{\vect{X}_{\cdot c\cdot\cdot} - \hat{\mu}_{ c}}{\hat{\sigma}_{c}}\big)^2\Big\}= 1
\end{equation}
Then, $\vect{\dot{X}}$ is reshaped as $\vect{\dot{X}}\in\mathds{R}^{B\times G\times C/G\times H\times W}$ to have $G$ groups of channels.
Next, $\forall g, b$,
\begin{equation}\label{eq:bcngn}
    \vect{\dot{Y}}_{bg\cdot\cdot\cdot} = \gamma_g^c\dfrac{\vect{\dot{X}}_{bg\cdot\cdot\cdot} - \mu_{bg\cdot\cdot\cdot}}{\sigma{_{bg\cdot\cdot\cdot}}}+ \beta_g^c
\end{equation}
Finally, $\vect{\dot{Y}}$ is reshaped back to $\vect{Y}\in\mathds{R}^{B\times C\times H\times W}$, which is the output of the Batch-Channel Normalization.

\subsection{Large- and Micro-batch Implementations}
Note that in Eq.~\ref{eq:bcnbn} and \ref{eq:bcngn}, only two statistics need batch information: $\hat{\mu}_{c}$ and $\hat{\sigma}_{c}$, as their values depend on more than one sample.
Depending on how we obtain the values of $\hat{\mu}_{c}$ and $\hat{\sigma}_{c}$, we have different implementations for large-batch and micro-batch training settings.

\vspace{-0.15in}
\paragraph{Large-batch training.}
When the batch size is large, estimating $\hat{\mu}_{c}$ and $\hat{\sigma}_{c}$ is easy: we just use a Batch Normalization layer to achieve the function of Eq.~\ref{eq:bcnbn} and \ref{eq:bcn_purpose}.
As a result, the proposed BCN can be written as
\begin{equation}
    \text{BCN}(\vect{X})=\text{CN}(\text{BN}(\vect{X}))
\end{equation}
Implementing it is also easy with modern deep learning libraries, which is omitted here.

\vspace{-0.15in}
\paragraph{Micro-batch training.}
One of the motivations of channel normalization is to allow deep networks to train on tasks where the batch size is limited by the GPU memory.
Therefore, it is important for Batch-Channel Normalization to be able to work in the micro-batch training setting.

\begin{algorithm}[t]
\SetAlgoLined
\SetKwInput{KwInput}{Input}
\SetKwInput{KwOutput}{Output}
 \KwInput{$\vect{X}\in\mathds{R}^{B\times C\times H\times W}$, the current estimates of $\hat{\mu}_c$ and $\hat{\sigma}^2_c$, and the update rate $r$.}
 \KwOutput{Normalized $\vect{Y}$.}
 Compute $\dot{\mu}_{c}\leftarrow\frac{1}{BHW}\sum_{b,h,w}\vect{X}_{b,c,h,w}$\;
 Compute $\dot{\sigma}^2_{c}\leftarrow\frac{1}{BHW}\sum_{b,h,w} \big( \vect{X}_{b,c,h,w} - \hat{\mu}_c \big)^2$\;
 Update $\hat{\mu}_c\leftarrow \hat{\mu}_c + r (\dot{\mu}_c - \hat{\mu}_c)$\;
 Update $\hat{\sigma}^2_c\leftarrow \hat{\sigma}^2_c + r (\dot{\sigma}^2_c - \hat{\sigma}^2_c)$\;
 Normalize $\vect{\dot{X}}_{\cdot c\cdot\cdot}=\gamma_c^b \dfrac{\vect{X}_{\cdot c\cdot\cdot} - \hat{\mu}_{ c}}{\hat{\sigma}_{c}}+\beta_{c}^b$\;
 Reshape $\vect{\dot{X}}$ to $\vect{\dot{X}}\in\mathds{R}^{B\times G\times C/G\times H\times W}$\;
 Normalize $\vect{\dot{Y}}_{bg\cdot\cdot\cdot} = \gamma_g^c\dfrac{\vect{\dot{X}}_{bg\cdot\cdot\cdot} - \mu_{bg\cdot\cdot\cdot}}{\sigma{_{bg\cdot\cdot\cdot}}}+ \beta_g^c$\;
 Reshape $\vect{\dot{Y}}$ to $\vect{{Y}}\in\mathds{R}^{B\times C\times H\times W}$\;
 \caption{Micro-batch BCN}\label{alg:1}
\end{algorithm}

Algorithm~\ref{alg:1} shows the feed-forwarding implementation of the micro-batch Batch-Channel Normalization.
The basic idea behind this algorithm is to constantly estimate the values of $\hat{\mu}_c$ and $\hat{\sigma}_c$, which are initialized as $0$ and $1$, respectively, and normalize $\vect{X}$ based on these estimates.
It is worth noting that in the algorithm, $\hat{\mu}_c$ and $\hat{\sigma}_c$ are not updated by the gradients computed from the loss function;
instead, they are updated towards more accurate estimates of those statistics.
Step 3 and 4 in Algorithm~\ref{alg:1} resemble the update steps in gradient descent; thus, the implementation can also be written in gradient descent by storing the difference $\Delta\hat{\mu}_c$ and $\Delta\hat{\sigma}_c$ as their gradients.
Moreover, we set the update rate $r$ to be the learning rate of trainable parameters.

Algorithm~\ref{alg:1} also raises an interesting question: when researchers study the micro-batch issue of BN before, why not just use the estimates to batch-normalize the features?
In fact, \cite{batchrenorm} tries a similar idea, but does not fully solve the micro-batch issue: it needs a bootstrap phase to make the estimates meaningful, and the performances are usually not satisfactory.
The underlying difference between micro-batch BCN and \cite{batchrenorm} is that BCN has a channel normalization following the estimate-based normalization.
This makes the previously unstable estimate-based normalization stable, and the reduction of Lipschitz constants which speeds up training is also done in the channel-based normalization part, which is also impossible to do in estimate-based normalization.
In summary, \textit{channel-based normalization makes estimate-based normalization possible, and estimate-based normalization helps channel-based normalization to keep models away from elimination singularities}. 

\subsection{Is Batch-Channel Normalization Redundant?}
Batch- and channel-based normalizations are similar in many ways.
Is BCN thus redundant as it normalizes normalized features?
Our answer is \textbf{no}.
Channel normalizations need batch knowledge to keep the models away from elimination singularities; at the same time, it also brings benefits to the batch-based normalization, including:

{\noindent\bf Batch knowledge without large batches.}
Since BCN runs in both large-batch and micro-batch settings, it provides a way to utilize batch knowledge to normalize activations without relying on large training batch sizes.

{\noindent\bf Additional non-linearity.}
Batch Normalization is linear in the test mode or when the batch size is large in training.
By contrast, channel-based normalization methods, as they normalize each sample individually, are not linear.
They will add strong non-linearity and increase the model capacity.

{\noindent\bf Test-time normalization.}
Unlike BN that relies on estimated statistics on the training dataset for testing, channel normalization normalizes testing data again, thus allows the statistics to adapt to different samples.
As a result, channel normalization will be more robust to statistical changes and show better generalizability for unseen data. 

\section{Experimental Results}\label{sec:exp}
In this section, we test the proposed BCN in popular vision benchmarks, including image classification on CIFAR-10/100~\cite{cifar} and ImageNet~\cite{ILSVRC15}, semantic segmentation on PASCAL VOC 2012~\cite{pascal}, and object detection and instance segmentation on COCO~\cite{coco}.

\subsection{Image Classification on CIFAR}
CIFAR has two image datasets, CIFAR-10 (C10) and CIFAR-100 (C100).
Both C10 and C100 have color images of size $32\times 32$.
C10 dataset has 10 categories while C100 dataset has 100 categories.
Each of C10 and C100 has 50,000 images for training and 10,000 images for testing and the categories are balanced in terms of the number of samples.
In all the experiments shown here, the standard data augmentation schemes are used, \textit{i.e.}, mirroring and shifting, for these two datasets.
We also standardizes each channel of the datasets for data pre-processing.

Table~\ref{tab:cifar1} shows the experimental results that compare our proposed BCN with BN and GN.
The results are grouped into 4 parts based on whether the training is large-batch or micro-batch, and whether the dataset is C10 and C100.
On C10, our proposed BCN is better than BN on large-batch training, and is better than GN (with or without WS) which is specifically designed for micro-batch training.
Here, micro-batch training assumes the batch size is 1, and RN110 is the 110-layer ResNet~\cite{resnet} with basic block as the building block.
The number of groups here for GN is $\min\{32, (\text{the number of channels}) / 4\}$.

Table~\ref{tab:cifar2} shows comparisons with more recent normalization methods, Switchable Normalization (SN)~\cite{switchnorm} and Dynamic Normalization (DN)~\cite{dynamicnorm} which were tested for a variant of ResNet for CIFAR: ResNet-18.
To provide readers with direct comparisons, we also evaluate BCN on ResNet-18 with the group number set to $32$ for models that use GN.
Again, all the results are organized based on whether they are trained in the micro-batch setting.
Based on the results shown in Table~\ref{tab:cifar1} and \ref{tab:cifar2}, it is clear that BCN is able to outperform the baselines effortlessly in both large-batch and micro-batch training settings.

\begin{table}[]
\small
    \centering
    \begin{tabular}{|c||c|c|cc||c|}
        \hline
        Dataset & Model & Method & Micro-Batch & WS & Error \\\hline\hline
        C10 & RN110 & BN & \xmark & \xmark & 6.43 \\
        C10 & RN110 & BCN & \xmark & \cmark & \bf5.90 \\\hline\hline
        C10 & RN110 & GN & \cmark & \xmark & 7.45 \\
        C10 & RN110 & GN & \cmark & \cmark & 6.82 \\
        C10 & RN110 & BCN & \cmark & \cmark & \bf6.31 \\\hline\hline
        C100 & RN110 & BN & \xmark & \xmark & 28.86 \\
        C100 & RN110 & BCN & \xmark & \cmark & \bf28.36 \\\hline\hline
        C100 & RN110 & GN & \cmark & \xmark & 32.86 \\
        C100 & RN110 & GN & \cmark & \cmark & 29.49 \\
        C100 & RN110 & BCN & \cmark & \cmark & \bf28.28 \\
        \hline
    \end{tabular}
    \caption{Error rates of a 110-layer ResNet~\cite{resnet} on CIFAR-10/100~\cite{cifar} trained with BN~\cite{batchnorm}, GN~\cite{groupnorm} and our BCN.
    The results are grouped based on dataset and large/micro-batch training.
    Micro-batch assumes $1$ sample per batch while large-batch uses 128 samples in each batch.
    WS indicates whether WS~\cite{qiao2019weight} is used for weights.}
    \label{tab:cifar1}
\end{table}

\begin{table}[]
\small
    \centering
    \begin{tabular}{|c||c|c|c||c|}
        \hline
        Dataset & Model & Method & Micro-Batch & Error \\\hline\hline
        C10 & RN18 & BN & \xmark & 5.20 \\
        C10 & RN18 & SN & \xmark & 5.60 \\
        C10 & RN18 & DN & \xmark & 5.02 \\
        C10 & RN18 & BCN & \xmark & \bf 4.96 \\\hline\hline
        C10 & RN18 & BN & \cmark & 8.45 \\
        C10 & RN18 & SN & \cmark & 7.62 \\
        C10 & RN18 & DN & \cmark & 7.55 \\
        C10 & RN18 & BCN & \cmark & \bf5.43 \\
        \hline
    \end{tabular}
    \caption{Error rates of ResNet-18 on CIFAR-10 trained with SN~\cite{switchnorm}, DN~\cite{dynamicnorm} and our BCN.
    The results are grouped based on large/micro-batch training.
    The performances of BN, SN and DN are from \cite{dynamicnorm}.
    Micro-batch for BN, SN and DN uses 2 images per batch, while BCN uses 1.}
    \label{tab:cifar2}
\end{table}

\subsection{Image Classification on ImageNet}

\begin{figure}
    \centering
    \includegraphics[width=\linewidth]{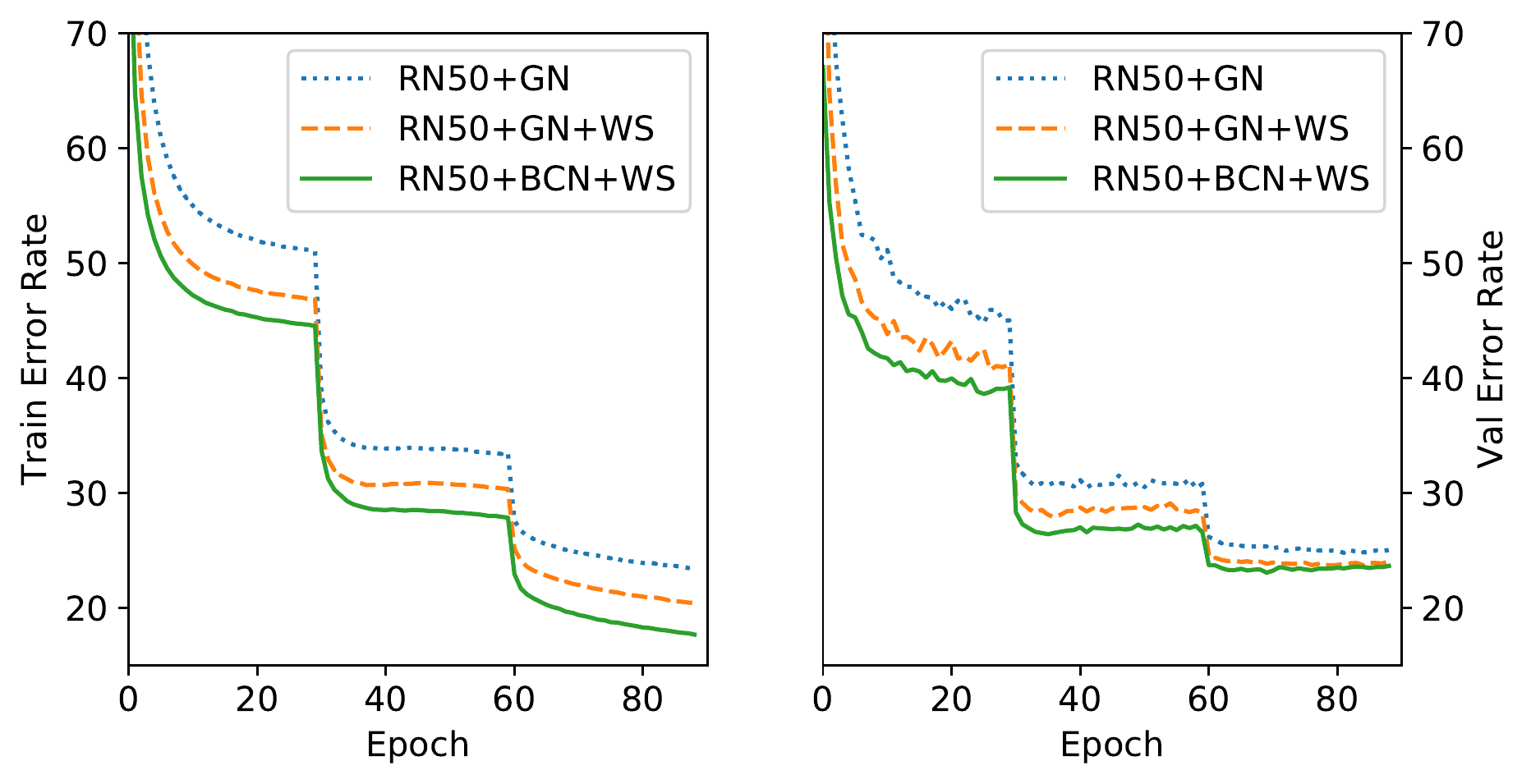}
    \caption{Training and validation error rates of ResNet-50 on ImageNet. The comparison is between the baselines GN ~\cite{groupnorm}, GN + WS~\cite{qiao2019weight}, and our proposed Batch-Channel Normalization (BCN) with WS. Our method BCN not only significantly improves the training speed, it also lowers the error rates of the final models by a comfortable margin.}
    \label{fig:imagenet}
\end{figure}

\begin{table}[]
\small
    \centering
    \begin{tabular}{|c||c|c|c||c|c|}
        \hline
        Dataset & Model & Method & WS & Top-1 & Top-5 \\\hline\hline
        ImageNet & RN50 & BN & \xmark & 24.30 & 7.19 \\
        ImageNet & RN50 & BN & \cmark & 23.76 & 7.13 \\
        ImageNet & RN50 & GN & \cmark & 23.72 & 6.99 \\
        ImageNet & RN50 & BCN & \cmark & \bf23.09 & \bf6.55 \\\hline\hline
        ImageNet & RN101 & BN & \xmark & 22.44 & 6.21 \\
        ImageNet & RN101 & BN & \cmark & 21.89 & 6.01 \\
        ImageNet & RN101 & GN & \cmark & 22.10 & 6.07 \\
        ImageNet & RN101 & BCN & \cmark & \bf 21.29 & \bf 5.60 \\\hline\hline
        ImageNet & RX50 & BN & \xmark & 22.60 & 6.29 \\
        ImageNet & RX50 & GN & \cmark & 22.71 & 6.38 \\
        ImageNet & RX50 & BCN & \cmark & \bf22.08 & \bf5.99 \\
        \hline
    \end{tabular}
    \caption{Top-1/5 error rates of ResNet-50, ResNet-101, and ResNeXt-50 on ImageNet. The test size is $224\times224$ with center cropping. All normalizations are trained with batch size $32$ or $64$ per GPU without synchronization.}
    \label{tab:imagenet}
\end{table}

\begin{table*}[]
\small
    \centering
    \begin{tabular}{|c|c|c||ccc||ccc||ccc||ccc|}
        \hline
        Model & Method & WS & AP$^b$ & AP$^b_{.5}$ & AP$^b_{.75}$ & AP$^b_{l}$ & AP$^b_{m}$ & AP$^b_{s}$ & AP$^m$ & AP$^m_{.5}$ & AP$^m_{.75}$ & AP$^m_{l}$ & AP$^m_{m}$ & AP$^m_{s}$\\\hline\hline
        RN50 & GN & \xmark & 39.8 & 60.5 & 43.4 & 52.4 & 42.9 & 23.0 & 36.1 & 57.4 & 38.7 & 53.6 & 38.6 & 16.9 \\
        RN50 & GN & \cmark & 40.8 & 61.6 & 44.8 & 52.7 & 44.0 & 23.5 & 36.5 & 58.5 & 38.9 & 53.5 & 39.3 & 16.6\\
        RN50 & BCN & \cmark & \bf 41.4 & \bf 62.2 & \bf 45.2 & \bf 54.7 & \bf 45.0 & \bf 24.2 & \bf 37.3 & \bf 59.4 & \bf 39.8 & \bf 55.0 & \bf 40.1 & \bf 17.9 \\\hline\hline
        RN101 & GN & \xmark & 41.5 & 62.0 & 45.5 & 54.8 &45.0 &24.1 & 37.0 &59.0 &39.6 & 54.5 &40.0 &17.5\\
        RN101 & GN & \cmark & 42.7 & 63.6 & 46.8 & 56.0 & 46.0 & 25.7 & 37.9 & 60.4 & 40.7 & 56.3 & 40.6 & 18.2 \\
        RN101 & BCN & \cmark & \bf43.6 & \bf64.4 & \bf47.9 & \bf57.4 & \bf47.5 & \bf25.6 & \bf 39.1 &\bf 61.4 &\bf 42.2 &\bf 57.3 &\bf 42.1 &\bf 19.1\\
        \hline
    \end{tabular}
    \caption{Object detection and instance segmentation results on COCO val2017~\cite{coco} of Mask R-CNN~\cite{maskrcnn} and FPN~\cite{FPN} with ResNet-50 and ResNet-101~\cite{resnet} as backbone.
    The models are trained with different normalization methods, which are used in their backbones, bounding box heads, and mask heads.}
    \label{tab:mask}
\end{table*}

This section shows the results of training models with BCN on ImageNet~\cite{ILSVRC15}.
The ImageNet dataset contains 1.28 million color images for training and 5,000 images for validation.
There are 1,000 categories in the datasets, which are roughly balanced.
We adopt the same training and testing procedures used in \cite{qiao2019weight}, and the baseline performances are copied from them.

Fig.~\ref{fig:imagenet} shows the training dynamics of ResNet-50 with GN, GN+WS and BCN+WS, and Table~\ref{tab:imagenet} shows the top-1 and top-5 error rates of ResNet-50, ResNet-101 and ResNeXt-50 trained with different normalization methods.
From the results, we observe that adding batch information to channel-based normalizations strongly improves their accuracy.
As a result, GN, whose performances are similar to BN when used with WS, now is able to achieve better results than the BN baselines.
And we find improvements not only in the final model accuracy, but also in the training speed.
As shown in Fig.~\ref{fig:imagenet}, we see a big drop of training error rates at each epoch.
This demonstrates that the model is now farther from elimination singularities, resulting in an easier and faster learning.

\subsection{Semantic Segmentation on PASCAL VOC}

\begin{table}[]
\small
    \centering
    \begin{tabular}{|c||c|c|c||c|}
        \hline
        Dataset & Model & Method & WS & mIoU \\\hline\hline
        VOC Val & RN101 & GN & \xmark & 74.90 \\
        VOC Val & RN101 & GN & \cmark & 77.20 \\
        VOC Val & RN101 & BN & \xmark & 76.49 \\
        VOC Val & RN101 & BN & \cmark & 77.15 \\
        VOC Val & RN101 & BCN & \cmark & \bf78.10 \\
        \hline
    \end{tabular}
    \caption{Comparisons of semantic segmentation performance of DeepLabV3~\cite{deeplabv3} trained with different normalizations on PASCAL VOC 2012~\cite{pascal} validation set. Output stride is 16, without multi-scale or flipping when testing.}
    \label{tab:voc}
\end{table}

After evaluating BCN on classification tasks, we test it on dense prediction tasks.
We start with semantic segmentation on PASCAL VOC~\cite{pascal}.
We choose DeepLabV3~\cite{deeplabv3} as the evaluation model for its good performances and its use of the pre-trained ResNet-101 backbone.

Table~\ref{tab:voc} shows our results on PASCAL VOC, which has $21$ different categories with background included.
We take the common practice to prepare the dataset, and the training set is augmented by
the annotations provided in \cite{pascalextra}, thus has 10,582 images.
We take our ResNet-101 pre-trained on ImageNet and finetune it for the task.
Here, we list all the implementation details for easy reproductions of our results:
the batch size is set to $16$, the image crop size is $513$, the learning rate follows polynomial decay with an initial rate $0.007$.
The model is trained for $30K$ iterations, and the multi-grid is $(1,1,1)$ instead of $(1, 2, 4)$.
For testing, the output stride is set to $16$, and we do not use multi-scale or horizontal flipping test augmentation.
As shown in Table~\ref{tab:voc}, by only changing the normalization methods from BN and GN to our BCN, mIoU increases by about $1\%$, which is a significant improvement for PASCAL VOC dataset.
As we strictly follow the hyper-parameters used in the previous work, there could be even more room of improvements if we tune them to favor BCN, which we do not explore in this paper and leave to future work.

\subsection{Object Detection and Segmentation on COCO}

\begin{table}[]
\small
\setlength{\tabcolsep}{0.35em}
    \centering
    \begin{tabular}{|c|c|c||ccc||ccc|}
        \hline
        Model & Method & WS & AP$^b$ & AP$^b_{.5}$ & AP$^b_{.75}$ & AP$^b_{l}$ & AP$^b_{m}$ & AP$^b_{s}$ \\\hline\hline
        RN50 & GN & \xmark & 38.0 & 59.1 & 41.2 & 49.5 &40.9 &22.4 \\
        RN50 & GN & \cmark & 38.9 & 60.4 & 42.1 & 50.4 &42.4 &23.5 \\
        RN50 & BCN & \cmark & \bf 39.7 & \bf 60.9 & \bf 43.1 & \bf 51.7 & \bf 43.2 & \bf 24.0 \\\hline\hline
        RN101 & GN & \xmark & 39.7 & 60.9 & 43.3 & 51.9 & 43.3 &23.1  \\
        RN101 & GN & \cmark & 41.3 & 62.8 & 45.1 & 53.9 & 45.2 & 24.7 \\
        RN101 & BCN & \cmark & \bf 41.8 & \bf 63.4 & \bf 45.8 & \bf 54.1 & \bf 45.6 & \bf 25.6 \\\hline\hline
        RX50 & GN & \cmark &  39.9 & 61.7 & 43.4 & 51.1 & 43.6 & 24.2 \\
        RX50 & BCN & \cmark & \bf40.5 & \bf62.2 & \bf44.2 & \bf52.3 & \bf44.3 & \bf25.1  \\
        \hline
    \end{tabular}
    \caption{Object detection results on COCO using Faster R-CNN~\cite{fasterrcnn} and
FPN with different normalization methods.}
    \label{tab:fast}
\end{table}

As we have introduced in Sec.~\ref{sec:norm}, our BCN can also be used for micro-batch training, which we will evaluate in this section by showing detection and segmentation results on COCO~\cite{coco}.
It is a very fundamental vision task yet has memory issues when large batch sizes are used.

We take our ResNet-50 and ResNet-101 normalized by BCN pre-trained on ImageNet as the starting point of the backbone, and fine-tune it on COCO train2017 dataset.
After training, the models are tested on COCO val2017 dataset.
We use 4 GPUs to train all the models, each GPU has one training sample.
Learning rate is configured according to the batch size following the common practice provided in \cite{mmdetection,Detectron2018}.
Specifically, we use 1X learning rate schedule for Faster R-CNN and 2X learning rate schedule for Mask R-CNN to get the results reported in this paper.
We use FPN~\cite{FPN} and the 4conv1fc bounding box head.
We add BCN to the backbone, bounding box heads, and mask heads.
We keep everything else untouched to maximize comparison fairness.
Please see \cite{mmdetection,Detectron2018} for more details.

Table~\ref{tab:mask} shows the results of Mask R-CNN~\cite{maskrcnn} between our BCN with GN and GN+WS, and Table~\ref{tab:fast} shows the comparisons on Faster R-CNN~\cite{fasterrcnn}.
The results shown in the tables are the \textit{Average Precision} for bounding box (AP$^b$) and instance segmentation (AP$^m$).
As the tables demonstrate, our BCN is able to outperform the baseline methods by a comfortable margin.

Experiments on COCO differ from the previous results on ImageNet and PASCAL VOC in that they train models in the micro-batch setting: each GPU can only have one training sample and the GPUs are not synchronized -- the batch size would be 4 even if they do, which is still not large.
The results on ImageNet and PASCAL VOC show that when large-batch training is available, having batch information will strongly improve the results.
And the experiments on COCO demonstrate that even when large-batch is not available, having an estimate-based batch normalization is also going to be helpful and will provide improvements.
The improvements over WS when GN is used show that although WS is able to alleviate the statistical difference issue, it does not fully solve it.
However, we do not just discard WS when we use BCN because WS still has the smoothing effect on the loss landscape which improves training from another perspective.
Overall, the results in this section prove the necessity of keeping models away from elimination singularities when training neural networks, and BCN improves results by avoiding them along the training trajectory.
\section{Related Work}\label{sec:related}
Deep neural networks advance state-of-the-arts in many computer vision tasks~\cite{deeplab,densenet,alexnet,fcnn,fewshot,qiao2018deep,qiu2017unrealcv,vggnet,sort,wang2018multi,yang2018knowledge,zhang2018single}.
But deep networks are hard to train.
To speed up training, proper model initializations are widely used as well as data normalization based on the assumption of the data distribution~\cite{glorot2010understanding,he2015delving}.
On top of data normalization and model initialization, Batch Normalization~\cite{batchnorm} is proposed to ensure certain distributions so that the normalization effects will not fade away during training.
By performing normalization along the batch dimension, Batch Normalization achieves state-of-the-art performances in many tasks in addition to accelerating the training process.
When the batch size decreases, however, the performances of Batch Normalization drop dramatically since the batch statistics are not representative enough of the dataset statistics.
Unlike Batch Normalization that works on the batch dimension, Layer Normalization~\cite{layernorm} normalizes data on the channel dimension, Instance Normalization~\cite{instnorm} does Batch Normalization for each sample individually.
Group Normalization~\cite{groupnorm} also normalizes features on the channel dimension, but it finds a better middle point between Layer Normalization and Instance Normalization.

Batch Normalization, Layer Normalization, Group Normalization, and Instance Normalization are all activation-based normalization methods.
Besides them, there are also weight-based normalization methods, such as Weight Normalization~\cite{weightnorm} and Weight Standardization~\cite{qiao2019weight,huang2017centered}.
Weight Normalization decouples the length and the direction of the weights, while Weight Standardization ensures the weights to have zero mean and unit variance.
Weight Standardization narrows the performance gap between Batch Normalization and Group Normalization, therefore, in this paper, we use Weight Standardization for our proposed method to get all the results.

In this paper, we study normalization methods and Elimination Singularity~\cite{orhan2017skip,wei2008dynamics}.
There are also other perspectives to understand normalization methods.
For example, from the perspective of training robustness, BN is able to make optimization trajectories more robust to parameter initialization~\cite{im2016empirical}.
\cite{whybnworks,qiao2019weight} show that normalizations are able to reduce the Lipschitz constants of the loss and the gradients, thus the training becomes easier and faster.
From the angle of model generalization, \cite{morcos2018importance} shows that Batch Normalization relies less on single directions of activations, thus has better generalization properties, and \cite{luo2018towards} studies the regularization effects of Batch Normalization.
\cite{kohler2018towards} also explores length-direction decoupling in BN and \cite{weightnorm}.
Other work also approaches normalizations from the gradient explosion issues~\cite{DBLP:journals/corr/abs-1902-08129} and learning rate tuning~\cite{DBLP:journals/corr/abs-1812-03981}.

Our method uses Batch Normalization and Group Normalization at the same time for one layer.
Some previous work also uses multiple normalizations or a combined version of normalizations for one layer.
For example, SN~\cite{switchnorm} computes BN, IN, and LN at the same time and uses AutoML~\cite{pnas} to determine how to combine them.
SSN~\cite{shao2019ssn} uses SparsestMax to get sparse SN.
DN~\cite{dynamicnorm} proposes a more flexible form to represent normalizations and finds better normalizations.
Unlike them, our method is based on analysis from the angle of elimination singularity instead of AutoML, and our normalizations are used together as a composite function rather than linearly adding up the normalization effects in a flat way.

\section{Conclusion}\label{sec:conc}
In this paper, we approach the normalization methods from the perspective of elimination singularities.
We study how different normalizations can keep their models away from the elimination singularities, since getting close to them will harm the training of the models.
We observe that Batch Normalization (BN) is able to guarantee a far distance from the elimination singularities, while Layer Normalization (LN) and Group Normalization (GN) are unable to keep far distances.
We also observe that the situation of LN is worse than that of GN, and Weight Standardization (WS) is able to alleviate this issue.
These findings are consistent with their performances.
We notice that the cause of LN and GN being unable to keep models away is their lack of batch knowledge.
Therefore, to improve their performances, we propose Batch-Channel Normalization (BCN), which adds batch knowledge to channel-normalized models.
BCN is able to run and improve the performances in both large-batch and micro-batch settings.
We test it on many popular vision benchmarks.
The experimental results show that it is able to outperform the baselines effortlessly.

{\small
\bibliographystyle{ieee_fullname}
\bibliography{egbib}
}

\end{document}